\documentclass[letterpaper]{article} 
\usepackage[preprint]{aaai2027}  
\usepackage[hyphens]{url}  
\usepackage{graphicx} 
\usepackage{natbib}  
\usepackage{caption} 
\usepackage{algorithm}
\usepackage{algorithmic}

\usepackage{newfloat}
\usepackage{listings}
\DeclareCaptionStyle{ruled}{labelfont=normalfont,labelsep=colon,strut=off} 
\floatstyle{ruled}
\newfloat{listing}{tb}{lst}{}
\floatname{listing}{Listing}

\usepackage{booktabs}
\usepackage{tabularx}
\usepackage{amsmath}

\title{ABSClaw: From Invented Worlds to Reality-Grounded Social Simulation}
\author{
    Shaobin He\textsuperscript{\rm 1,2},
    Yan Gao\textsuperscript{\rm 2,3}\corresponding,
    Quan Zheng\textsuperscript{\rm 2,4},
    Zhen Liu\textsuperscript{\rm 2,3},
    Jiyan He\textsuperscript{\rm 2,3},
    Shuxin Zheng\textsuperscript{\rm 2,3}
}

\affiliations{
    \textsuperscript{\rm 1}Tianjin University\\
    \textsuperscript{\rm 2}Zhongguancun Academy\\
    \textsuperscript{\rm 3}Zhongguancun Institute of Artificial Intelligence\\
    \textsuperscript{\rm 4}Beijing Normal University\\
    sb\_he@tju.edu.cn, gaoyan@zgci.ac.cn
}

\begin{document}

\maketitle

\begin{abstract}
LLM-based artificial societies exhibit rich behavior but often lack scientific grounding: personas and environments are assumed, interaction networks are synthetic, and simulators are generated as free-form code, making results difficult to tie to real populations and sensitive to the backbone model. We argue that simulated worlds should be \emph{sampled} from observed data, while simulation logic should be \emph{implemented} by composing operations from a shared, task-independent vocabulary. In this paper, we introduce \textsc{ABSClaw}, a zero-code agent that compiles natural-language requests into executable, reality-grounded social simulations. 
ABSClaw grounds \emph{who} acts, \emph{where} they act, and \emph{with whom} they interact in survey and census data, geospatial sources, and real networks.  It expresses state transitions through 26 atomic operations. To support an open and extensible ecosystem, ABSClaw further allows capabilities to be added as lifecycle-scoped skills without modifying its core.
Furthermore, we propose ABSBench, a benchmark of 193 real-world social-simulation tasks. On ABSBench, ABSClaw achieves the best overall performance among the evaluated social simulators. Three real-world case studies show that grounding personas, environments, and interaction networks each improves agreement with observed outcomes. With one runtime skill, \textsc{ABSClaw} executes 13 population-scale simulations, collectively involving 25 million agents. Together, these results demonstrate \textsc{ABSClaw}'s real-world fidelity, extensibility, and scalability.
\end{abstract}

\section{Introduction}
Large language model (LLM) agents are transforming agent-based modeling into a flexible framework for constructing interactive artificial societies. This shift opens new opportunities for exploratory social science, policy prototyping, and the study of social intelligence \citep{bonabeau2002agent,park2023generative,aher2023simulate}.


Yet behavioral richness is not scientific grounding. We identify three gaps that keep LLM-based simulations closer to \emph{toy societies} than to reality-grounded ones. \textbf{(i) Worlds are invented rather than grounded in observed data.} Most systems prompt personas, assume generic environments, and construct random or fully connected networks. Even data-driven systems usually ground only one component, leaving people, places, and social ties unmatched to a real population or setting. \textbf{(ii) Mechanism implementation is entangled with simulator construction.} To implement a specified social mechanism, existing systems often ask an LLM to generate the entire simulator as free-form code, including both task-specific logic and basic state-transition machinery. Coding errors and incidental implementation choices can therefore change the simulated dynamics, making results sensitive to the backbone model and difficult to reproduce or compare. \textbf{(iii) Capabilities are hard-wired rather than composable.} Different studies require different data sources, mechanisms, and analyses: public-health models may need demographic profiles, mobility models may need roads and points of interest, and opinion models may need survey attitudes and empirical networks. A fixed simulator cannot cover this long tail without repeated redesign, motivating a general-purpose, pluggable framework for social simulation.


We present \textsc{ABSClaw}, a zero-code agent for constructing reality-grounded social simulations. \textsc{ABSClaw} addresses the three gaps above by grounding personas, environments, and social networks in real-world sources; compiling simulation mechanisms into programs over a fixed vocabulary of 26 atomic operations rather than generating unconstrained code; and allowing new capabilities to be added as extensible skills without modifying the core simulator. 
Our main contributions are threefold:

\begin{figure*}[t]
    \centering
    \includegraphics[width=\textwidth]{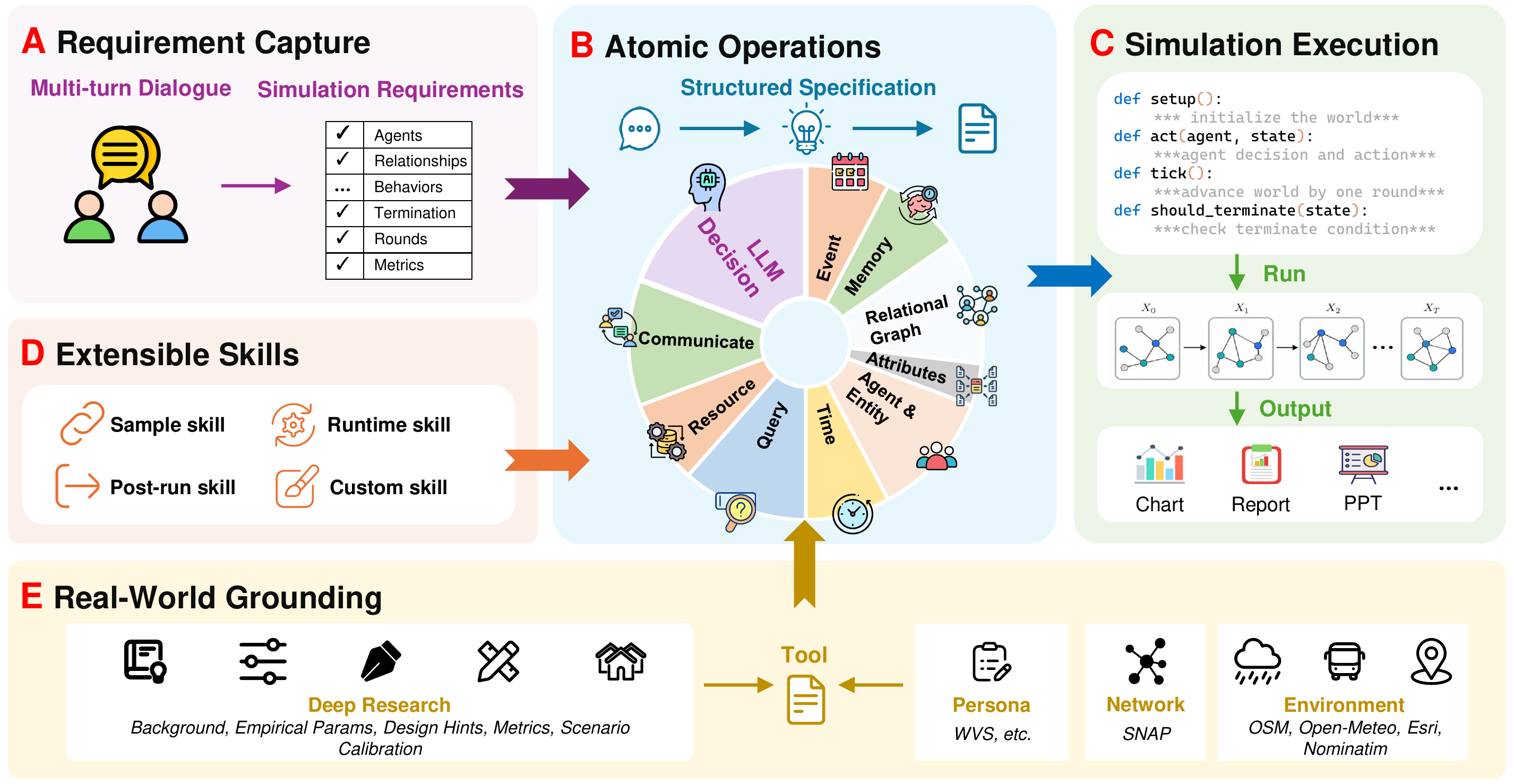}
    \caption{ABSClaw workflow.}
    \label{fig:overview}
\end{figure*}

\begin{itemize}
    \item \textbf{A zero-code agent for reality-grounded social simulation.}
     We introduce \textsc{ABSClaw}, which transforms natural-language requests
    into executable social simulations and jointly grounds simulated personas,
    environments, and social networks in real-world data.

    \item \textbf{A constrained and extensible mechanism-programming framework.}
    We define a fixed vocabulary of 26 atomic operations over a shared
    simulation runtime. Given a natural-language description, the LLM implements
    the requested simulation mechanism by composing these operations into an
    executable program, rather than generating the entire simulator as
    unconstrained code. We further enable new
    data-access, execution, and analysis capabilities to be added without modifying the core simulator.

    \item \textbf{The first benchmark and systematic evaluation for executable social-simulation construction.}
To the best of our knowledge, we introduce \textsc{ABSBench}, the first benchmark for evaluating the construction of executable social simulations from natural-language requests, comprising 193 real-world tasks across ten domains. Systematic evaluations show that \textsc{ABSClaw} achieves the best overall performance, reduces cross-backbone performance variation by 61.6\%, improves real-world fidelity through grounded personas, environments, and networks, and scales to 25 million agents with a single runtime skill.
\end{itemize}

\section{Related Work}

\paragraph{Generative social simulation.}
Agent-based modeling explains macro-level phenomena through individual rules and interactions \citep{bonabeau2002agent}, while LLMs support language-mediated perception and decision-making without fully hand-coded behavior \citep{ghaffarzadegan2024generative}. Surveys span individual-, scenario-, and society-level simulation, identifying persistent challenges in validation, scalability, reproducibility, and connecting individual behavior to collective outcomes \citep{mou2026individual,orlando2026mapping}. Generative Agents introduced memory, reflection, and planning, while Concordia provided reusable abstractions for agents, environments, and game masters \citep{park2023generative,vezhnevets2023concordia}. Later systems extend task-oriented interaction and team formation \citep{li2025metaagents}, general-purpose simulator generation \citep{tang2025gensim}, realistic urban and socioeconomic environments \citep{piao2025agentsociety}, and large-scale social-media simulation \citep{yang2024oasis}. Other work models persistent affect, identity adaptation, real-time event-driven activity, and graph-optimized hybrid LLM--ABM execution \citep{fu2026sentipolis,fan2026aivilization,allegrini2026botverse,zhou2026gasim}; Light Society combines full LLMs with distilled surrogates to exceed one billion agents \citep{guan2025earthscale}. Despite advances in cognition, interaction, coverage, and scale, these systems do not systematically combine real-world grounding with a stable, task-independent simulation representation.

\paragraph{Empirical grounding and behavioral validity.}
Persona conditioning can reproduce demographic response patterns \citep{argyle2023one,aher2023simulate,park2024selfreports}. However, SocioBench reports only about 30--40\% accuracy across more than 480,000 survey records, with substantial variation across domains and demographic groups \citep{wang2025sociobench}; moreover, consistent self-reports may coexist with declining persona expression over long interactions \citep{gonnermannmueller2026stable}. Plausible role-playing therefore does not guarantee valid collective behavior, which also depends on environmental dynamics, exposure, scheduling, and initialization \citep{li2026agentsalone}. System-level approaches ground simulations in observed networks and environments: S$^3$ uses real social-network data \citep{gao2023s3}; SocioVerse aligns users, environments, scenarios, and behaviors using ten million user records \citep{zhang2025socioverse}. These systems demonstrate the value of empirical grounding but typically embed specific data sources and alignment procedures in domain-specific platforms.

\paragraph{Automated simulator construction and research automation.}
YuLan-OneSim generates executable social simulations from natural-language dialogue \citep{wang2025onesim}; S-Researcher automates experiment design, participant simulation, analysis, and reporting \citep{wang2026socialscientists}; and SOCIA-EVO combines evidence blueprints with structural evolution and parameter calibration \citep{hua2026socia}. Although these systems reduce manual effort, they largely generate or revise unconstrained code, making state mutation, scheduling, and interaction semantics difficult to inspect and compare across tasks or backbone models. Related work improves reliability through constrained generation or by separating adaptive reasoning from deterministic execution: PICARD rejects grammar-invalid decoding steps \citep{scholak2021picard}; Synchromesh imposes syntactic and semantic completion constraints \citep{poesia2022synchromesh}. ABSClaw applies the same principle to simulator construction by compiling requested mechanisms into bounded control flow over a fixed vocabulary of auditable state-transition operations, rather than regenerating infrastructure for each task.

\section{Overview}
\label{sec:overview}

We formulate open-ended social simulation as a compilation problem.
Given a natural-language request $x$, optional persona, environment, and
network sources $D=(D_p,D_e,D_n)$, and a catalog of installed skills
$\mathcal{S}$, ABSClaw generates an executable lifecycle program
$P=(P_{\mathrm{setup}},P_{\mathrm{act}},P_{\mathrm{tick}},
P_{\mathrm{term}})$.

As shown in Figure~\ref{fig:overview}, ABSClaw proceeds in four steps.
First, it resolves the user's multi-turn request into a structured simulation
specification $s$. Second, it materializes the required source data as
run-scoped typed records and exposes only a compact manifest to the code
compiler. Third, it compiles $s$, the manifest, and selected skill interfaces
into a program over a fixed vocabulary of atomic operations $\mathcal{O}$.
Finally, the executor advances the simulated world and records execution
traces, metrics, and output artifacts.

\section{ABSClaw}
\label{sec:absclaw}
\subsection{Source-Conditioned World Materialization}
\label{sec:materialization}

ABSClaw separates source-specific data processing from simulation-code
generation. Let $c$ denote the source configuration,
$\Theta=A(x,c)$ the resolved provider configuration, and
$K=\operatorname{Catalog}(D;\Theta)$ a compact catalog of available fields
and observed values. The compilation pipeline is
\begin{equation}
\begin{aligned}
(s,F) &= C_{\mathrm{spec}}(x,K),\\
(B,H) &= M(D;\Theta,F),\\
P &= C_{\mathrm{code}}(s,H),\\
W_0 &= \operatorname{Setup}(P,B;\mathcal{O}),
\end{aligned}
\label{eq:pipeline}
\end{equation}
where $F$ contains semantic filtering requirements,
$B=B_p\cup B_e\cup B_n$ is a typed sample bank, and $H$ is its compact
manifest. The code compiler receives only $s$ and $H$.

For each source axis $a\in\{p,e,n\}$, the materialization backend normalizes
a source snapshot and resolves the requested filters against values observed
in that snapshot:
\begin{equation}
\begin{aligned}
\widehat F_a
  &= \operatorname{Resolve}\!\left(F_a,
     \operatorname{Obs}(N_a(D_a))\right),\\
B_a &= \operatorname{Take}_{L_a}
      \left\{r:q_{\widehat F_a}(r)\right\}.
\end{aligned}
\label{eq:materialize}
\end{equation}
Materialization is deterministic given a fixed source snapshot, provider
configuration, and resolved specification.

\paragraph{Persona.}
Persona records contain sample keys and typed attributes. ABSClaw materializes them from individual-level surveys such as the World Values Survey (WVS) \citep{haerpfer2022wvs} and census-derived demographic data.

\paragraph{Network.}
Network records contain endpoint keys, relation layers, and edge attributes
parsed from SNAP topologies \citep{leskovec2014snap}. When network and persona
sources are combined, a deterministic partial mapping projects the anonymous
topology onto sampled persona keys.

\paragraph{Environment.}
Environment records represent non-acting entities constructed from place, radius, and time constraints. ABSClaw resolves locations through Nominatim and Esri, obtains road and point-of-interest context from OpenStreetMap, and retrieves time-specific weather conditions from Open-Meteo \citep{haklay2008osm,nominatim2026,esri2026,openmeteo2023}.

\paragraph{Deep Research.} 
 ABSClaw can conduct broad web research before formal source sampling. It decomposes the user request into a multi-angle search plan spanning conceptual background, theory, statistics, policy, stakeholders, behavioral mechanisms, and evaluation metrics.

\subsection{Atomic-Operation Intermediate Representation}
\label{sec:atomic-ir}
ABSClaw targets the language $\mathcal{L}(\mathcal{O},\mathcal{S}_x)$, which combines bounded
Python control flow, a fixed operation set $\mathcal{O}$, and exports from the
selected skill subset $\mathcal{S}_x$.

The operation vocabulary was developed in two stages. We first reviewed
recurring simulator semantics in foundational agent-based modeling and
social-simulation literature
\citep{bonabeau2002agent,gilbert2005simulation,macal2010tutorial,grimm2020odd}.
The review covered entities and state variables, initialization, scheduling,
interaction, communication, resource exchange, and measurement. Domain experts
then consolidated overlapping candidates and identified missing primitives.
This process yielded 26 operations in ten functional categories; the complete
vocabulary is listed in Table~\ref{tab:absclaw-atomic-ops}. Each operation acts on a core world state $W$ as a typed partial state
transformer:
\begin{equation}
o_\theta:(W,a)\mapsto(W',y,\ell),
\label{eq:operation}
\end{equation}
where $a$ is the argument tuple, $y$ is the return value, and $\ell$ is an
effect record appended to the execution trace. Predicates, loops, and
arithmetic remain explicit in the generated program. Among these operations, llm\_decide requests a runtime model decision. Deterministic mechanisms are expressed
with ordinary program expressions and therefore do not invoke this operation. Stage~2 emits the lifecycle functions setup, act, tick, and should\_terminate. After abstract-syntax-tree and
interface validation, the assembler links them into an executable module.

\begin{table}[tb]
  \centering
  {
  \setlength{\tabcolsep}{2.5pt}
  \renewcommand{\arraystretch}{1.08}
  \begin{tabularx}{\columnwidth}{
    @{}>{\raggedright\arraybackslash}p{0.28\columnwidth}
       >{\raggedright\arraybackslash}X@{}
  }
    \toprule
    Category & Atomic operations \\
    \midrule
    Attributes &
      \texttt{set\_attr}, \texttt{get} \\
    Entity lifecycle &
      \texttt{create\_agent}, \texttt{create\_entity} \\
    Relations &
      \texttt{create\_edge}, \texttt{set\_edge}, \texttt{get\_edge},
      \texttt{get\_neighbors} \\
    Resources &
      \texttt{transfer} \\
    Communication &
      \texttt{send\_message}, \texttt{read\_messages},
      \texttt{delete\_message} \\
    Memory &
      \texttt{add\_memory}, \texttt{retrieve\_memory} \\
    Events &
      \texttt{fire\_event}, \texttt{schedule\_event},
      \texttt{cancel\_event} \\
    Time/control &
      \texttt{get\_time}, \texttt{terminate\_simulation} \\
    Query and observation &
      \texttt{query\_entities}, \texttt{aggregate},
      \texttt{log\_metric}, \texttt{log\_decision},
      \texttt{query\_metric\_history}, \texttt{query\_metric\_values} \\
    LLM decision &
      \texttt{llm\_decide} \\
    \bottomrule
  \end{tabularx}
  }
  \caption{The 26 atomic operations in ABSClaw.}
  \label{tab:absclaw-atomic-ops}
\end{table}

\subsection{Lifecycle-Scoped Skills}
Sample Skills expose a compact pre-Stage~1 data catalog, allowing the specification model to select fields and values available from configured sources. Runtime Skills provide data accessors or deterministic helpers, and Post-run Skills analyze final artifacts. This enables drop-in extensibility without changing the core simulator or its state-transition semantics.

\begin{figure*}[t]
    \centering
    \includegraphics[width=\textwidth]{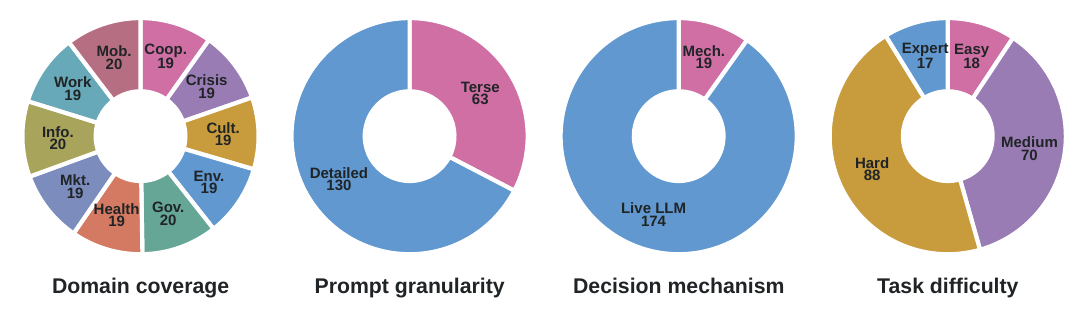}
    \caption{Composition of the 193 ABSBench tasks by (A) application domain, (B) prompt granularity, (C) decision mechanism, and (D) task difficulty. Segment angle represents task count. Domain labels denote Cooperation and public goods (Coop.), Crisis and conflict (Crisis), Culture and norms (Cult.), Environment and resources (Env.), Governance and institutions (Gov.), Health behavior (Health), Markets and reputation (Mkt.), Opinion and information (Info.), Organizations and work (Work), and Urban mobility (Mob.). In panel (C), Mech. denotes mechanical or offline decision policies.}
    \label{fig:absbench-composition}
\end{figure*}

\section{ABSBench}
\label{sec:absbench}

We introduce ABSBench, a benchmark for evaluating the compilation of natural-language requests into executable agent-based social simulations. ABSBench was constructed independently of the ABSClaw implementation. The benchmark contains 193 tasks curated to represent real-world social processes and practical modeling needs. Each task centers on interactions among heterogeneous actors, temporal changes in individual or collective states, and measurable system-level outcomes.


The benchmark was constructed in four stages. First, we defined ten domains to cover diverse interaction mechanisms and application settings. Second, we collected candidate modeling needs from academic literature and related resources. Third, domain experts removed tasks that were duplicate, underspecified, infeasible to evaluate, or not meaningfully agent-based, while checking the remaining set for diversity and coverage. Fourth, an LLM standardized the retained seeds into a common task format, including actors, interactions, state transitions, temporal requirements, and outcome measures. Experts then verified each standardized task against its original seed and corrected any semantic drift or unsupported additions.

Every ABSBench task includes a public natural-language prompt, an evaluator-only behavioral specification, and structured metadata. Candidate programs run through a common interface, and their outputs are normalized into a final summary, per-round metric trajectories, and agent-indexed decision records. Validity (\(V\); 0--100) comprises base execution and required artifacts, task-constraint compliance, pipeline and semantic evidence, and artifact quality. Task correctness (\(T\); 0--100) is assigned by an LLM judge across implementation fidelity (accuracy), code quality, and result plausibility. For validation, two domain experts independently and blindly scored a stratified subset of 59 runs; mean judge--expert quadratic-weighted Cohen's \(\kappa=\mathrm{0.87}\) indicated strong agreement.

\paragraph{Domain coverage.}
Figure~\ref{fig:absbench-composition} shows near-uniform representation across ten domains. Every domain contributes between 9.8\% and 10.4\% of the benchmark.

\paragraph{Prompt granularity.}
ABSBench contains 130 detailed prompts (67.4\%) and 63 terse prompts (32.6\%). Detailed prompts range from 298 to 545 whitespace-delimited words. They specify the agent population, individual states, interaction rules, interventions, and required artifacts. Terse prompts range from 13 to 20 words. They state the modeling problem and intended outcome while leaving agent design and implementation choices to the evaluated system.

\paragraph{Decision mechanisms and difficulty.}
Figure~\ref{fig:absbench-composition} shows that 174 tasks (90.2\%) require live LLM calls for substantive agent decisions; the remaining 19 (9.8\%) use mechanical or offline policies. The task-difficulty chart reports 18 Easy tasks (9.3\%), 70 Medium tasks (36.3\%), 88 Hard tasks (45.6\%), and 17 Expert tasks (8.8\%). These labels summarize combined technical demands from agent heterogeneity, persistent state, interaction topology, temporal feedback, counterfactual comparisons, scalability, and output verification.

\section{Results}

\subsection{External System Comparison} 
We compared ABSClaw with SOCIA-EVO and OneSim because all three support end-to-end construction of executable simulators from general natural-language requests. Other influential platforms, including SocioVerse and AgentSociety, rely on dedicated environments, data infrastructures, or platform-specific workflows and are therefore not directly comparable under the ABSBench protocol.

\begin{figure}[!tb]
  \centering
  \includegraphics[width=\columnwidth]{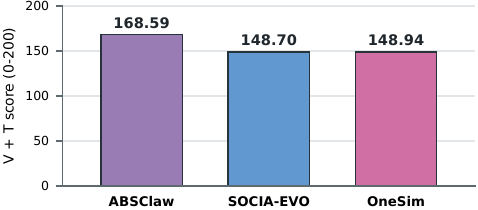}
  \caption{Overall external-system performance on all 193 ABSBench tasks. Bars show fixed-denominator mean \(V+T\) scores (0--200), with failed or missing outputs scored as zero. SOCIA-EVO and OneSim were adapted to the common ABSBench input/output interface while preserving their native simulator-construction mechanisms; all systems received semantically equivalent task information, and outputs were normalized to the same evaluation schema. GPT-5.5 was used for simulator construction and Gemini~2.5~Flash for runtime agent decisions in all three systems.} 
  \label{fig:external-system-comparison}
\end{figure}

\begin{figure}[!tb]
  \centering
  \includegraphics[width=\columnwidth]{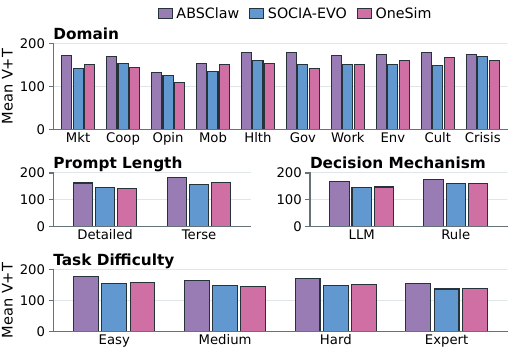}
  \caption{External-system performance by benchmark stratum. Bars show
fixed-denominator mean \(V+T\) scores (0--200) by application domain,
prompt granularity, decision mechanism, and difficulty; failed or missing outputs are scored as zero.}
  \label{fig:external-system-comparison-stratified}
\end{figure}

ABSClaw achieved 168.59 out of 200, exceeding SOCIA-EVO (148.70) by 19.89 points and OneSim (148.94) by 19.65 points (Figure~\ref{fig:external-system-comparison}). ABSClaw achieved the best performance across domains, prompt lengths, decision mechanisms, and task difficulties (Figure~\ref{fig:external-system-comparison-stratified}).

\subsection{Atomic Operations Reduce Backbone Sensitivity}

Across five generators, \emph{ABSClaw} outperformed the matched \emph{Spec-to-Code} baseline. Its $V{+}T$ scores ranged from 142.69 to 168.59 out of 200, compared with 69.90 to 137.36 for direct generation. The resulting gains ranged from $+29.19$ to $+72.80$ points, with a cross-generator mean of $+48.94$ points.

Across generators, Validity increased by 9.36--40.06 points and Task correctness by 15.44--32.74 points. Among 965 matched model--task evaluations, ABSClaw yielded successful executions with available scores in 931 cases (96.5\%), compared with 738 (76.5\%) for Spec-to-Code.

Atomic-operation assembly also reduced sensitivity to generator choice. The between-generator range in $V{+}T$ contracted from 67.47 points under direct generation to 25.89 points with ABSClaw, a 61.6\% reduction. Within the Qwen family, replacing Qwen3.5-122B-A10B with Qwen3.5-27B increased ABSClaw performance by 10.60 points but decreased Spec-to-Code performance by 17.21 points. This yielded a $+27.80$-point interaction (paired task-bootstrap 95\% CI $[16.36,39.37]$). Across the Claude models, Haiku retained 86.8\% of Opus performance under ABSClaw but only 51.7\% under direct generation. The corresponding interaction was $+43.61$ points (95\% CI $[29.61,57.65]$).

\begin{table}[t]
\centering
{\small
\setlength{\tabcolsep}{1mm}
\begin{tabular}{@{}lrrrrrrr@{}}
\toprule
& \multicolumn{3}{c}{ABSClaw} & \multicolumn{3}{c}{Spec-to-Code} & \\
\cmidrule(lr){2-4}\cmidrule(lr){5-7}
& $V$ & $T$ & $V{+}T$ & $V$ & $T$ & $V{+}T$
& \multicolumn{1}{c}{\shortstack{$\Delta$\\95\% CI}} \\
\midrule
Opus 4.8
& 92.8 & 71.5 & 164.3 & 79.1 & 56.0 & 135.1
& \shortstack[r]{+29.2\\{[20.4,38.0]}} \\
Qwen-A10B
& 91.6 & 53.6 & 145.2 & 67.4 & 36.0 & 103.4
& \shortstack[r]{+41.8\\{[33.0,50.9]}} \\
Qwen-27B
& 96.8 & 59.1 & 155.9 & 57.5 & 28.7 & 86.2
& \shortstack[r]{+69.6\\{[60.4,79.1]}} \\
GPT-5.5
& 90.1 & 78.5 & \textbf{168.6} & 80.7 & 56.6 & 137.3
& \shortstack[r]{+31.2\\{[23.8,38.6]}} \\
Haiku 4.5
& 86.9 & 55.8 & 142.7 & 46.8 & 23.1 & 69.9
& \shortstack[r]{\textbf{+72.8}\\{\textbf{[62.3,83.3]}}} \\
\bottomrule
\end{tabular}
}
\caption{Matched comparison of ABSClaw and Spec-to-Code on 193 tasks under identical prompts, generators, runtime settings, and evaluation procedures. Scores use the fixed 193-task denominator, with failed or missing outputs assigned zero. $\Delta$ is the paired difference in $V{+}T$, computed before rounding; brackets show 95\% task-level percentile-bootstrap confidence intervals.}
\label{tab:main}
\end{table}

\begin{figure}[!tb]
  \centering
  \includegraphics[width=\columnwidth]{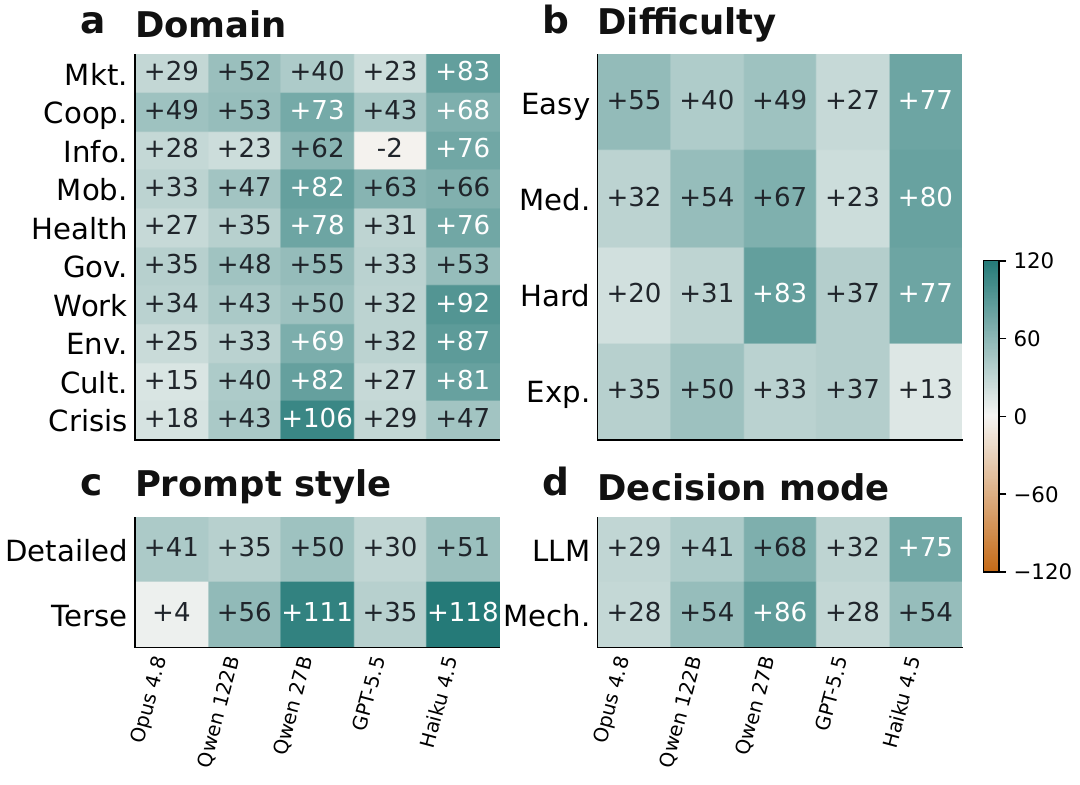}
  \caption{Stratified \(V+T\) difference between ABSClaw and Spec-to-Code by (a) application domain, (b) task difficulty, (c) prompt style, and (d) decision mode. Positive values favor ABSClaw. Each task--method--model combination contributes one run.}
  \label{fig:absbench-stratified-advantage}
\end{figure}

ABSClaw achieved higher $V{+}T$ scores in 844 of 965 matched model--task evaluations (87.5\%), with 101 reversals and 20 ties. The advantage of ABSClaw was also consistent across benchmark strata (Figure~\ref{fig:absbench-stratified-advantage}). It outperformed Spec-to-Code in all 20 difficulty--model, 10 prompt-style--model, and 10 decision-mode--model comparisons, and in 49 of 50 domain--model comparisons. The robustness benefit was especially pronounced on the 63 terse or ambiguous tasks. Under ABSClaw, Haiku scored 140.17 compared with 148.71 for Opus, retaining 94.3\% of Opus performance. Under Spec-to-Code, Haiku scored 21.70 compared with 144.79 for Opus, retaining only 15.0\%. This produced a $+114.56$-point interaction (paired task-bootstrap 95\% CI $[91.10,138.00]$). This pattern is consistent with the operation-constrained intermediate representation supplying shared simulator structure when task descriptions leave that structure implicit.

\subsection{Three-Axis Real-World Alignment}
We selected three real-world cases from ABSBench to evaluate ABSClaw across complementary alignment mechanisms: persona alignment grounds \emph{who} acts, environment alignment grounds the local conditions under which agents act, and network alignment grounds \emph{with whom} they interact.

To rule out direct external retrieval and minimize information leakage, we disabled ABSClaw's Deep Research module in all three studies. 

\begin{figure}[!tb]\centering\includegraphics[width=\columnwidth]{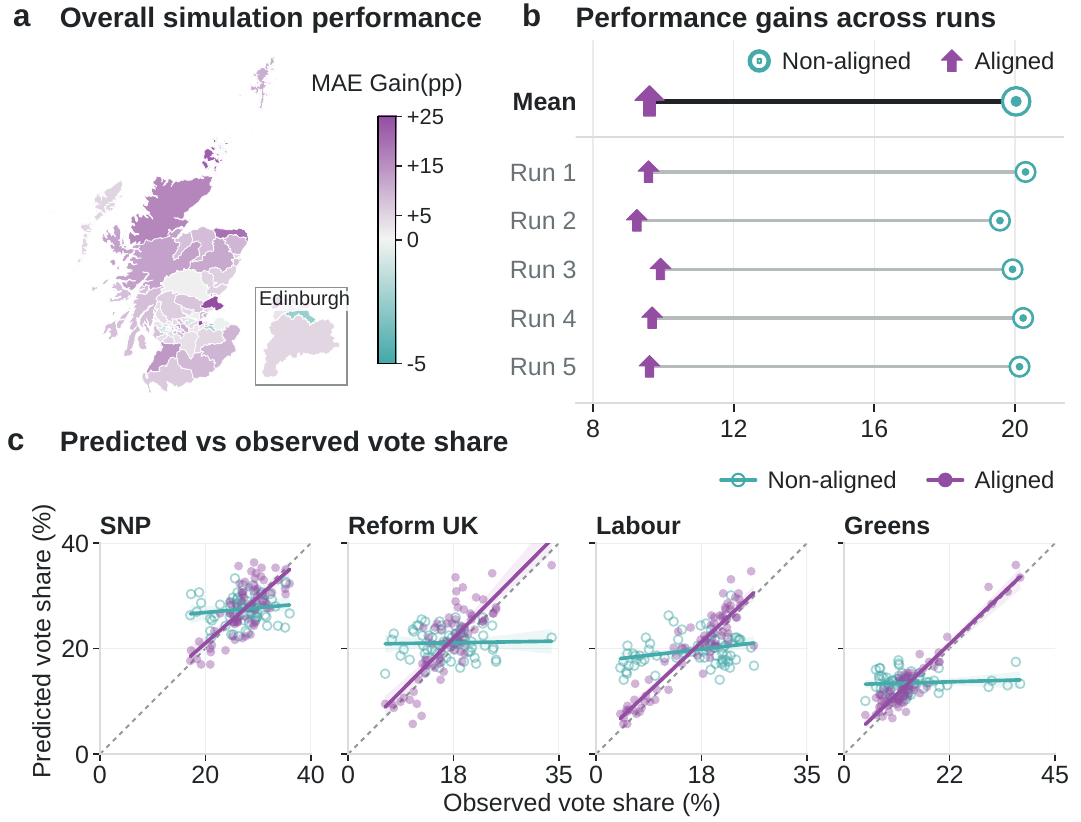}\caption{Effect of persona alignment on the 2026 Scottish Parliament election simulation: (a) constituency-level reductions in party-vote total-variation error, (b) paired errors across five seeds, and (c) observed versus simulated party vote shares. In the election study, the real-world event took place on 7 May 2026, after the published knowledge cutoffs of GPT-5.5 (1 December 2025) and Gemini 2.5 Flash (January 2025).}\label{fig:persona}\end{figure}

\paragraph{Persona alignment.}
We first modeled regional-list voting in the 2026 Scottish Parliament election across 73 constituencies (Fig.~\ref{fig:persona}). The aligned condition retained Census-grounded persona records in their source constituencies, whereas the non-aligned condition randomly permuted the identical persona pool across constituencies. Across five paired seeds, with 60 agents for every constituency, condition, and seed (43,800 decisions), the pooled equal-constituency Party-TV error fell from \(16.90\) to \(9.05\)~pp. This corresponds to a \(46.4\%\) relative reduction and an absolute improvement of \(7.85\)~pp (95\% CI: \(6.48\)--\(9.27\)~pp). Alignment improved 68 of 73 constituencies and all five seed-level comparisons, demonstrating that persona alignment substantially improves the geographic calibration of simulated electoral behavior.

\begin{figure}[!tb]\centering\includegraphics[width=\columnwidth]{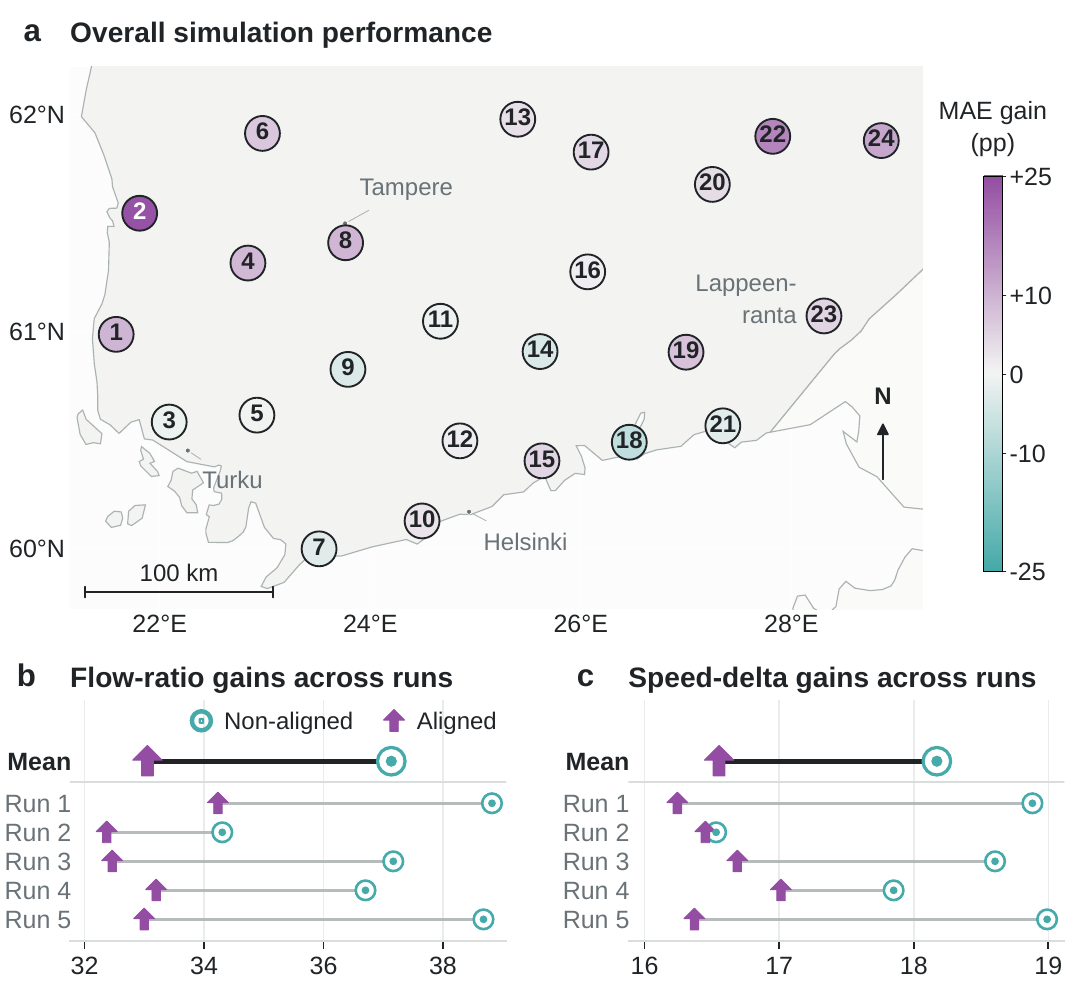}\caption{Effect of environment alignment during Storm Valtteri: (a) Site-level flow-ratio MAE gains; (b,c) paired flow-ratio and speed-delta MAEs across five seeds. In the storm and diffusion studies, prompts were masked to expose only decision-relevant observations, such as road and weather conditions, while withholding place names, event identities, and other revealing cues.}\label{fig:environment}\end{figure}

\paragraph{Environment alignment.}
We next reconstructed road-traffic disruption during Storm Valtteri at 24 observation sites in southern Finland (Fig.~\ref{fig:environment}). The aligned condition provided site-specific road and land-cover attributes together with local weather at each round; the non-aligned condition repeated one spatially pooled environment and weather sequence across all sites. Alignment reduced flow-ratio MAE from \(37.14\) to \(33.05\)~pp, an improvement of \(4.09\)~pp (95\% CI: \(0.64\)--\(7.56\)~pp), and reduced speed-delta MAE from \(18.17\) to \(16.55~\mathrm{km\,h^{-1}}\), an improvement of \(1.62~\mathrm{km\,h^{-1}}\) (95\% CI: \(-0.60\)--\(3.92~\mathrm{km\,h^{-1}}\)). Both errors decreased in all five seeds.

\begin{figure}[!tb]\centering\includegraphics[width=\columnwidth]{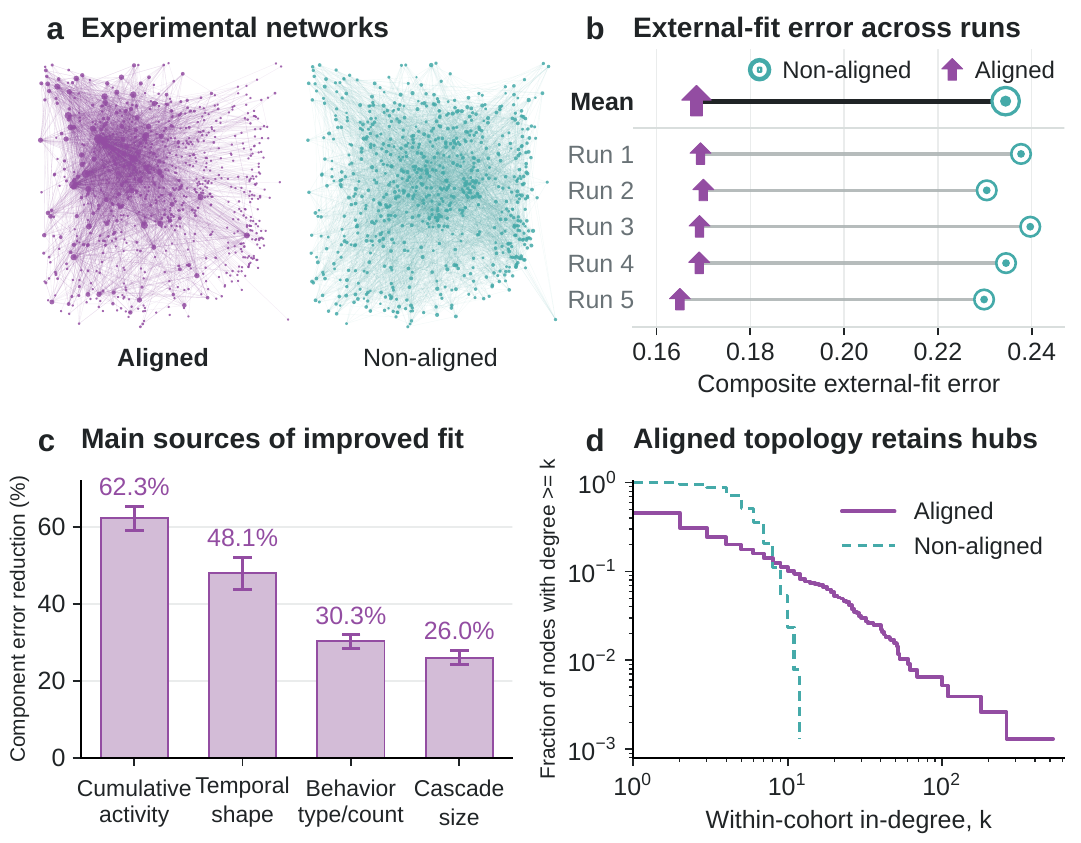}\caption{Effect of network alignment on Higgs information diffusion: (a) Empirical and randomized interaction networks; (b) composite external-fit error by seed and mean; (c) error reduction by outcome component; and (d) in-degree complementary cumulative distributions.}\label{fig:network}\end{figure}

\paragraph{Network alignment.}Finally, we evaluated information diffusion following CERN's Higgs-boson announcement (Fig.~\ref{fig:network}). The aligned condition instantiated the empirical Higgs interaction network, whereas the non-aligned condition used a randomized directed network with the same 768 agents, 3,701 edges, and overall density. Across ten rounds and five paired seeds (76,800 agent-round decisions), alignment reduced the mean composite external-fit error from \(0.2344\) to \(0.1685\), a \(28.12\%\) reduction (95\% CI: \(27.08\)--\(28.99\%\)), with lower error in all five seeds. The empirical interaction network therefore enables ABSClaw to recover real-world diffusion dynamics substantially more faithfully.

\subsection{Scaling ABSClaw through a runtime skill}
A central goal of ABSClaw is to support a unified ecosystem in which social simulation capabilities can be added through skills. To evaluate this design, we integrated \emph{Adaptive Prototype Simulation} (APS) as a runtime skill. APS is a framework for scaling multi-round LLM-agent social simulations while substantially reducing the required online model calls~\cite{zheng2026aps}. Through the existing skill mechanism, the integration equips ABSClaw with population-scale social simulation capability.

Using the ABSBench task and evaluation format, we constructed 13 held-out population-scale tasks, including three with real-world grounding (Table~\ref{tab:absclaw-aps-scale-13}). Tasks contained 0.5--5.0 million agents, totaling 25 million.

\begin{table}[!tb]
\centering
\begin{tabularx}{\columnwidth}{@{}Xrr@{}}
\toprule
Held-out scenario & Agents & $V{+}T$ \\
\midrule
Disaster insurance and institutional trust
    & 1.5M & 167 \\
Agricultural pest-response coordination
    & 1.0M & 158 \\
Open-science practice adoption
    & 0.5M & 175 \\
Remote work and regional migration
    & 3.0M & 183 \\
Colorectal-screening uptake
    & 2.0M & 149 \\
Public-benefit access and take-up
    & 2.5M & 162 \\
AI automation and job transition
    & 1.5M & 173 \\
Heatwave electricity-demand response
    & 5.0M & 157 \\
Family eldercare norms and burden
    & 1.0M & 156 \\
Wildfire warning and evacuation response
    & 4.0M & 171 \\
\midrule
Barcelona metro-area festival attendance
    & 1.0M & 167 \\
Greater Tokyo airport--rail choice
    & 1.0M & 194 \\
Cultural diffusion on empirical topology
    & 1.0M & 180 \\
\midrule
\textbf{Mean / total}
    & \textbf{25.0M}
    & \textbf{168.6} \\
\bottomrule
\end{tabularx}
\caption{ABSClaw+APS results on 13 population-scale tasks. 
GPT-5.5 served as the ABSClaw generation model, Gemini~2.5~Flash served as the runtime agent model within APS, and Claude~Opus~4.8 served as the evaluator.}
\label{tab:absclaw-aps-scale-13}
\end{table}

All 13 runs passed the validity evaluation with $V=100$, and a mean total score of 168.6/200 (Table~\ref{tab:absclaw-aps-scale-13}). The three real-world-grounded tasks achieved a mean total score of 180.3/200. These results show that ABSClaw can acquire and reuse population-scale social simulation capability by adding APS as a runtime skill, supporting its role as a unified skill ecosystem for agent-based social simulation.

\section{Conclusion}
We introduced ABSClaw, a zero-code framework for compiling natural-language requests into executable, reality-grounded social simulations. ABSClaw grounds personas, environments, and interaction networks in observed data, while expressing simulation logic through 26 shared atomic operations and extensible lifecycle-scoped skills. On the 193-task ABSBench, it achieved the best overall performance among evaluated systems and reduced cross-backbone variation by 61.6\% compared with unconstrained Spec-to-Code generation. Three real-world case studies showed that grounding each component improves agreement with observed outcomes, and a single APS runtime skill enabled 13 population-scale simulations totaling 25 million agents. These results demonstrate that shared, auditable simulation operations can improve reliability, composability, and scalability across diverse social-simulation tasks.

\bibliography{references}

\end{document}